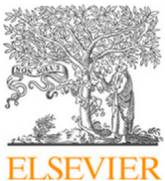
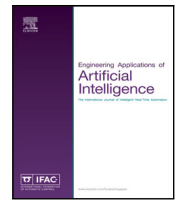

Contents lists available at ScienceDirect

# Engineering Applications of Artificial Intelligence

journal homepage: www.elsevier.com/locate/engappai

Research paper

# Power transformer health index and life span assessment: A comprehensive review of conventional and machine learning based approaches

Syeda Tahreem Zahra [a], Syed Kashif Imdad [a], Sohail Khan [c], Sohail Khalid [b], Nauman Anwar Baig [d,*]

[a] *Department of Electrical Engineering, HITEC University, Taxila, Pakistan*
[b] *Department of Electrical Engineering, Riphah International University, Islamabad, Pakistan*
[c] *Sino-Pak Centre for Artificial Intelligence, Pak-Austria Fachhochschule: Institute of Applied Sciences and Technology, Pakistan*
[d] *School of Computing, Engineering and Technology, Robert Gordon University, United Kingdom*

## ARTICLE INFO

*Keywords:*
Power transformers
Health index
Remaining life span
Machine learning
Dissolved gas analysis
Fiber optic sensors
Frequency domain spectroscopy
Artificial intelligence
Support vector machines
Random forest
K-nearest neighbor
Artificial neural networks
Genetic algorithm
Particle swarm optimization

## ABSTRACT

Power transformers play a critical role within the electrical power system, making their health assessment and the prediction of their remaining lifespan paramount for the purpose of ensuring efficient operation and facilitating effective maintenance planning. This paper undertakes a comprehensive examination of existent literature, with a primary focus on both conventional and cutting-edge techniques employed within this domain. The merits and demerits of recent methodologies and techniques are subjected to meticulous scrutiny and explication. Furthermore, this paper expounds upon intelligent fault diagnosis methodologies and delves into the most widely utilized intelligent algorithms for the assessment of transformer conditions. Diverse Artificial Intelligence (AI) approaches, including Artificial Neural Networks (ANN) and Convolutional Neural Network (CNN), Support Vector Machine (SVM), Random Forest (RF), Genetic Algorithm (GA), and Particle Swarm Optimization (PSO), are elucidated offering pragmatic solutions for enhancing the performance of transformer fault diagnosis. The amalgamation of multiple AI methodologies and the exploration of time-series analysis further contribute to the augmentation of diagnostic precision and the early detection of faults in transformers. By furnishing a comprehensive panorama of AI applications in the field of transformer fault diagnosis, this study lays the groundwork for future research endeavors and the progression of this critical area of study.

## 1. Introduction

The Asset management plays a crucial role in optimizing the utilization and minimizing costs in the electrical energy industry. This strategy aims to maximize the lifespan of existing equipment, including power transformers, to achieve a higher return on investment. Power transformers are essential components in power systems, enabling the efficient transmission and distribution of electricity and known for their reliability and potential lifespan of up to 60 years (Zhou et al., 2021; Fei and Zhang, 2009). The reliable operation of these transformers is paramount for maintaining the stability and functionality of the entire power network. However, transformers are subjected to aging, degradation, and potential failures over time and these unexpected failures can lead to significant losses for utilities and consumers, with lengthy repair or replacement duration exacerbating economic losses and risks. Therefore, accurate assessment of their health condition and prediction of remaining life span are essential for proactive maintenance and replacement strategies.

To mitigate these risks, effective asset management strategies are necessary to maintain power transformers in optimal operating conditions, minimizing the chances of power outages. Such strategies encompass two main components: life assessment and decision options based on economic considerations. These economic-based programs encompass various activities throughout the transformer's life cycle, including design, construction, operation, maintenance, repair, upgrading, replacement, and disposal (Tjernberg, 2018). Condition Monitoring (CM) systems play a vital role in asset management by facilitating early detection of potential faults during the operation and maintenance phases. CM systems utilize specialized equipment for monitoring and employ data analysis techniques to predict trends and assess the current performance of the monitored equipment. Hence, asset management strategies, supported by comprehensive condition monitoring systems, are essential for ensuring the optimal operation and longevity of power transformers. By implementing proactive maintenance measures, utilities can minimize the risks associated with transformer failures, prevent power outages, and optimize resource allocation and costs.

* Corresponding author.
  *E-mail address:* n.baig@rgu.ac.uk (N.A. Baig).

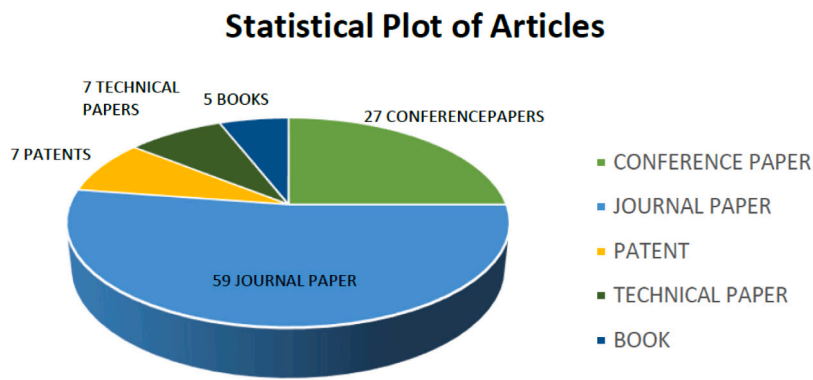

**Fig. 1.** Statistical plot of article types focusing on transformer health assessment.

The energy industry has long acknowledged the significance of incorporating CM into power equipment, a recognition dating back to the early 1990s. The availability of accurate and dependable diagnostic tools assumes paramount importance in ensuring the success of CM endeavors. There is a growing demand for non-invasive diagnostic and monitoring techniques to assess the health of power equipment. Current research endeavors aim to establish a platform that harnesses sensor data and equipment-specific information for this purpose. Power utilities are proactively seeking intelligent analytical tools that can increase the reliability through the deployment of automated fault indicators (Balamurugan and Ananthanarayanan, 2018).

The incorporation of Artificial Intelligence (AI) techniques, such as Machine Learning (ML) and data analytics, holds immense potential for CM systems, enabling more precise fault detection, condition evaluation, and the formulation of predictive maintenance strategies (Khalil, 2018; Islam et al., 2017). Among these methods, supervised ML approaches have emerged as effective tools for power transformer health assessment and life span prediction. These techniques utilize historical data and relevant parameters to develop predictive models capable of estimating the health index of transformers and forecasting their remaining useful life. By leveraging advanced algorithms and statistical analysis, ML models can analyze large volumes of data and identify intricate patterns that indicate transformer degradation and potential failure. The prediction of power transformer health index and remaining life span offers several benefits to power utilities and maintenance practitioners. It facilitates early detection of developing faults, allowing timely interventions to prevent catastrophic failures and reduce downtime and associated costs (Taha et al., 2021). Moreover, accurate life span prediction optimizes maintenance strategies, enabling proactive replacement or refurbishment of transformers before failure occurs, thus avoiding unplanned outages and minimizing overall maintenance expenses. Furthermore, it aids in asset management and resource allocation, ensuring the efficient utilization of available resources.

By conducting a thorough examination of the existing literature, this review paper consolidates the current state-of-art in transformer health assessment and prognosis using supervised ML techniques. It offers valuable insights into the advancements made in this field, providing researchers, practitioners, and power system operators with a comprehensive understanding of the potential and challenges associated with these approaches. Furthermore, this review paper have critically analyzed and compared the performance of different supervised ML algorithms in power transformer health assessment and life span prediction. It considers studies that utilize various data sources, such as historical operational data, Dissolved Gas Analysis (DGA) results, temperature measurements, and other relevant parameters. The evaluation encompasses metrics such as accuracy, precision, recall, and F1-score to assess the effectiveness of the predictive models.

Figs. 1 and 2 gives a statistical plot and yearly distribution of all the articles referred in this review paper.

The findings and discussions presented in this review paper contributes to the body of knowledge on power transformer health assessment and life span prediction. It will help researchers and practitioners to understand the strengths and limitations of different supervised ML approaches in this context, as well as identify potential avenues for further research and improvement.

The next sections of the paper are organized as follows. Section 2 discusses the diagnostic techniques such as DGA, Frequency Domain Spectroscopy (FDS), Frequency Response Analysis (FRA), Total Acid Number (TAN) tests, and furanic compound analysis for assessing power transformer health and mentions ML's role in enhancing these methods. Section 3 discusses the methods that estimate the remaining lifespan of transformers by considering various aging factors, such as temperature, moisture, oxygen, mechanical stresses, and corrosive elements, which are interconnected and complex to characterize accurately, aiming to improve asset management decisions in transformer diagnostics. Section 4 discusses the ML models, including regression and classification employed to predict transformer oil or paper insulation aging properties based on sensor data, enabling proactive maintenance decisions, with regression models mapping aging properties, and classification models categorizing severity levels, and various ML techniques are utilized for transformer health assessment and fault detection, each offering unique advantages and applications. Author's opinion and Future work is discussed in Section 5 suggesting the application of ML to power transformers addressing challenges such as limited failure data, dataset diversity, transformer variations, imbalanced datasets, real-time fault detection, on-field condition assessment intricacies, and localization capabilities for transformer health assessment. The paper ends with conclusion in Section 6.

## 2. Power transformer health assessment

According to the existing literature, the well-being of a transformer predominantly hinges on the state of its oil-paper insulation (Zhang and Gockenbach, 2008; Abu-Siada and Islam, 2012; Jalbert et al., 2012). The practice of assessing transformer oil samples, as proposed in Zhang and Gockenbach (2008), holds more benefits than examining other transformer elements (like turns ratio, winding resistance, leakage reactance, etc.) when it comes to identifying faults and foreseeing the transformer's operational lifespan. In order to gauge the overall health status, the Health Index (HI) is employed. This index combines operational observations, on-site inspections, and laboratory tests to assist in managing transformers as valuable assets, a concept detailed by Jahromi et al. (2009) and Naderian et al. (2008). The insulation state significantly affects the HI calculation, particularly in situations with limited data concerning the transformer's service history and design. A comprehensive evaluation of transformer oil samples through electrical, physical and chemical tests in a laboratory setting is imperative to comprehend the insulation condition. Doing this process is known as scoring weighing method and this method involves assigning scores to each factor involve in transformer health index by the expert personnel (Zhang and Gockenbach, 2008; Jahromi et al., 2009; Naderian et al., 2008).

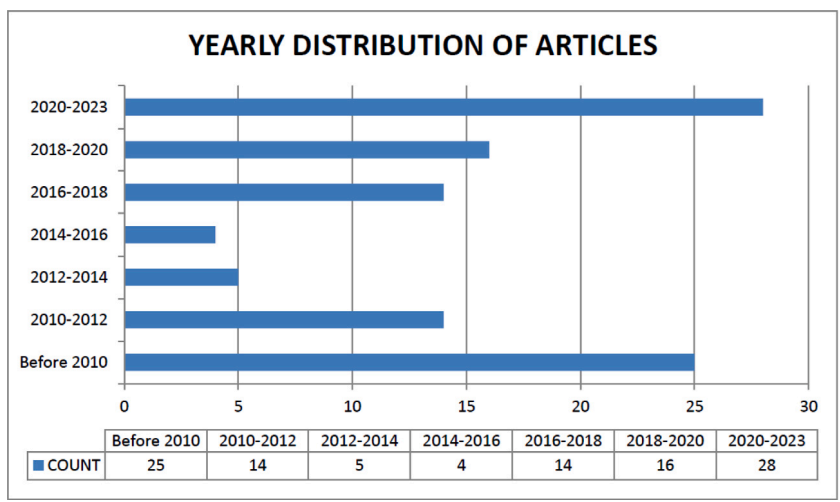

**Fig. 2.** Yearly distribution of articles featuring on transformer health studies.

**Table 1**
The scoring weighing paper condition (Rediansyah et al., 2021).

| Paper condition factor | Score | | | | | Weight |
|---|---|---|---|---|---|---|
| | 1 | 2 | 3 | 4 | 5 | |
| CO/CO2 | A | B | C | D | E | 0.171 |
| AGE(Years) | <20 | 20–30 | 30–40 | 40–60 | > 60 | 0.234 |
| 2FAL (ppb) | <100 | 100–500 | 500–1000 | 1000–5000 | > 5000 | 0.214 |

The HI computation involves conducting oil insulation laboratory tests in three major categories: Oil Quality Assessment (OQA), DGA, and Furfur Aldehyde (FFA) or furanic compounds tests (Jahromi et al., 2009; Naderian et al., 2008). DGA tests are performed to detect transformer internal faults, and they analyze gases like hydrogen (H2), ethane(C2H6), methane(CH4), ethyne(C2H2), ethene(C2H4), carbon dioxide (CO2) and carbon monoxide(CO). OQA tests determine the oil quality by measuring parameters like acidity, Breakdown Voltage (BDV), Interfacial Tension (IFT), water content, color and Dielectric Dissipation Factor (DDF) (IEEE, 2016). FFA tests, on the other hand, determines the extent to which the paper insulation degrades by measuring the furfur aldehyde (furan) content in transformer oil paper insulation, which indicates the aging of paper insulation (Rediansyah et al., 2021). The results from FFA tests are used to compute the HI using various formulas developed by Transformer Asset Management (TAM) field experts with the method described in Jahromi et al. (2009) and Naderian et al. (2008) being commonly used in publications. An example scoring Weighing for Oil Paper Insulation on the basis of formulas obtained by TAM field experts is given in Table 1.

In Guo and Guo (2022) the HI of a transformer is calculated by taking into account its aging, operational data, and on-site test results. Aging is evaluated not only by the transformer's operational duration but also by its designed lifespan. Operational data encompasses factors such as loading and pollution levels, which provide insight into the transformer's basic condition and are crucial for condition assessment. On-site test data offers a snapshot of the transformer's current state, making it essential for comprehensive condition evaluation.

### 2.1. Dissolved gas analysis

DGA is a type of technique used to characterize transformer oil and identify potential defects at an early stage (Perrier et al., 2012). This method is highly beneficial for preliminary aging detection and localization in transformers. During the operation of transformers, gases are produced resulting in decomposition of transformer oil, with the main gases being hydrogen gas, carbon monoxide and hydrocarbon. These gases are collected and analyzed to determine their presence and percentages. Severity of aging can be assessed based on the types and concentrations of gases detected. DGA involves two different stages. The first stage is extraction stage, where the dissolved gases are quantified. The second stage is the fault diagnosis stage, where the individual component gases are identified, leading to the detection of faults (Perrier et al., 2012). A scoring weighing of the DGA data analyzed by TAM field expert to assign scores to various gases present in insulation is given in Table 2.

**Table 2**
The scoring weighing for DGA (Rediansyah et al., 2021).

| The Level Scoring for DGA | | | | | |
|---|---|---|---|---|---|
| | H2 | CH4 | C2H6 | C2H2 | C2H4 |
| L1 | <80 | <90 | <90 | <1 | <50 |
| L2 | 80–200 | 90–150 | 90–170 | 1–2 | 50–100 |
| L3 | 200–320 | 150–210 | 170–250 | 2–3 | 100–150 |
| L4 | > 320 | > 210 | > 250 | > 3 | > 150 |
| The Rate Scoring for DGA | | | | | |
| | H2 | CH4 | C2H6 | C2H2 | C2H4 |
| R1 | <20 | <20 | <29 | 0 | <7 |
| R2 | 20–31 | 20–37 | 29–58 | – | 7–16 |
| R3 | 31–59 | 37–72 | 58–145 | 0 | 16–48 |
| R4 | > 59 | > 72 | > 145 | > 0 | > 48 |

To standardize the process of gas sampling from oil-filled instruments and specify various methods and tools for sampling and labeling (stage 1), the IEC 60567 standard is applied (Equipment, 2005). This standard ensures consistent and reliable sampling procedures. Additionally, the IEC 60599 standard address the interpretation of dissolved gases by the use of basic gas ratio (IEC, 2008). This standard provides guidance on how to interpret the various concentrations of free or dissolved gases in oil-filled electrical instruments to diagnose its condition and recommend appropriate actions. It can be used for equipment filled with insulating oils and insulated with cellulose paper or pressboards made from solid insulation. Although specific equipment types like transformers, reactors, bushings, and switch-gear are mentioned, the standard may be cautiously applied to other liquid–solid insulating systems as well.

**Table 3**
Comparative table for different methods in terms of performance matrices (Demirci et al., 2023).

| Method | Accuracy (%) | Precision (%) | Recall (%) | F-1 Score (%) |
|---|---|---|---|---|
| Traditional Classification | 85 | 82 | 83 | 82.5 |
| AI-Based Classification | 88 | 85 | 86 | 85.5 |
| Machine Learning + Sensor Fusion | 90 | 87 | 89 | 88 |
| Sequential Kalman Filter | 92 | 90 | 91 | 90.5 |
| Majority Voting Fusion Method | 91 | 89 | 90 | 89.5 |
| Dempster Shafer Evidence Theory | 93 | 91 | 92 | 91.5 |

Numerous methods for interpreting DGA data in oil filled power transformers have been proposed and employed, including the Doernenburg ratio, Rogers ratio, Logarithmic Nomograph, IEC basic ratios, Duval triangle, Key gases, and pentagon (Mehta et al., 2013; Lin and Yu, 2014). However, fault diagnosis using these methods relies on human experts' decisions and has various limitations. For example, the key gas method exhibits poor efficiency and accuracy for recognition, while the three-ratio methods suffer from inadequate coding. The shortcomings describe above hinder the detection of hidden and unusual faults in power transformers, and there is also the possibility of misjudging different types of defects with the same characteristics of gas.

To address these limitations, researchers have explored the application of ML based methods to use the DGA approach for detection of transformer failure. ML algorithms offer significant advantages, including reducing reliance on personnel expertise, improving consistency, and enabling the assessment of a large number of transformers with a vast amount of data (Bacha et al., 2012). One such example is the work by Bacha et al. who developed an intelligent technique for the classification of fault using ML for power transformer DGA (Bacha et al., 2012). Another new method uses Empirical Mode Decomposition (EMD) has been introduced in Sami and Bhuiyan (2020) to detect faults in transformers using DGA data. This method ranks DGA parameters based on their skewness and derives optimal sets of Intrinsic Mode Function (IMF) coefficients from these ranked parameters. The performance of this method surpasses both traditional and several existing machine learning techniques. Again a novel diagnostic technique utilizing Intrinsic Time-Scale Decomposition (ITD) has been created by Sami and Bhuiyan (2022) to identify faults in power transformers. This method ranks DGA parameters based on skewness and then extracts ITD-based features. An XGBoost classifier is used to select the optimal feature set and carry out the classification. It also showed even better F1-scores compared to previously discussed EMD-based technique. Another method presented in Demirci et al. (2023) combines gas data classified using machine learning with sensor fusion techniques to enhance diagnostic accuracy. It is found that using the Sequential Kalman filter, which is applied differently from previous studies, improves estimation accuracy to over 90%. This improvement is validated using the Majority Voting and Dempster Shafer Evidence Theory fusion methods, along with results from the IEC-TC-10 dataset. A comparative table of the performance matrices used in ML algorithms to determine overall performance of any methodology is given in Table 3.

Current research on equipment for DGA in transformer oil highlights several advanced technologies designed for real-time, on-site monitoring. These technologies are critical for ensuring the health and stability of power transformers by detecting potential faults early. In this context Tunable Diode Laser Absorption Spectroscopy (TDLAS) to detect multiple gas components in transformer oil are introduced. It involves an oil-gas separation system coupled with an optical detection system to analyze gases such as methane, ethylene, ethane, acetylene, carbon monoxide, and carbon dioxide. This setup allows for precise, real-time monitoring and has shown high accuracy and compliance with industry standards (Chen et al., 2021). N'cho and Fofana (2020) have introduced innovative sensors based on fiber optic to design and detect dissolved gases, such as hydrogen gas ($H_2$), carbon monoxide (CO), methane ($CH_4$) and acetylene ($C_2H_2$) in transformer oil, which serve as indicators of aging. These sensors offer the advantage of being involved with data acquisition for online health monitoring to took place online, eliminating the need for offline DGA techniques. Another method involves on-site chromatography analysis in which portable gas chromatography units separate and quantify different gas components directly from the transformer oil, providing immediate insights into the transformer's condition. Such systems are particularly useful for remote or hard-to-access transformers. Another innovative approach employs optical fiber sensors integrated into the transformer's structure. The fiber optic sensors detect changes in gas concentrations through variations in light transmission properties, offering a robust solution for transformer monitoring. Products like Serveron's on-line DGA monitors (e.g., Calisto 2, 5, and 9) offer comprehensive gas analysis capabilities. These devices continuously monitor gas levels in transformer oil and use advanced machine learning algorithms to diagnose potential faults, providing an early warning system to prevent failures (Bustamante et al., 2019).

Almost all of these and many other online devices and methods requires significant computational resources and expertise in machine learning. Hence, ML serves as a powerful tool to tackle challenges such as limited sampling, nonlinearity, and high-dimensional data, which are common in transformer DGA.

### 2.2. Partial discharge

One significant indicator of insulation weakness is the presence of Partial discharge (PD) activity (Barrios et al., 2019). The existence of PD indicates the deterioration of transformer insulation. PD can be characterized and categorized based on the type of defect or fault responsible for its occurrence and its location within the transformer. When PD occurs, positive and negative charges are neutralized, accompanied by a steep current pulse and radiating electromagnetic waves. Then the spectrum characteristics of electromagnetic radiation from PD are related to the geometry of the PD source and the insulation strength of the discharge gap. If the discharge gap is small or the insulation intensity is high, the steepness of the current pulse will be large and the radiation electromagnetic wave ability will be strong. The insulating strength of oil paper insulation in the transformer is high (Xuewei and Hanshan, 2019). Various factors influence the type of PD, such as pulse amplitude, occurrence time on the mains cycle (point on the wave at which it occurs), interval between discharges and number of discharges per second (Spurgeon et al., 2005; Ilkhechi et al., 2019). PD exhibits several effects, including electrical, acoustic, light emission, and electromagnetic disturbance, among others. Consequently, various methods and different sensors are employed for PD detection. For instance, a type of piezoelectric sensors is used for detection of PD via acoustic effects, while Optical Fiber based sensors can be used to detect the Partial Discharge signal. Additionally, Ultra high Frequency based sensors offer a newer approach to PD monitoring compared to traditional methods (Ilkhechi and Samimi, 2021; Naderi et al., 2007).

However, the information collected from PD can be vast and complex, making it challenging for humans to fully interpret and analyze it. To address this challenge, AI is employed to construct comprehensive frameworks that facilitate the discovery of various PD-related issues (Foo and Ghosh, 2002). Specifically, ML approaches have garnered significant interest from both industry and academia in intelligent PD diagnostics. ML models can obtain hierarchy based characteristics extracted from input data, leading to much more accurate and more reliable outputs, thus holding great potential for further development (Danikas et al., 2003).

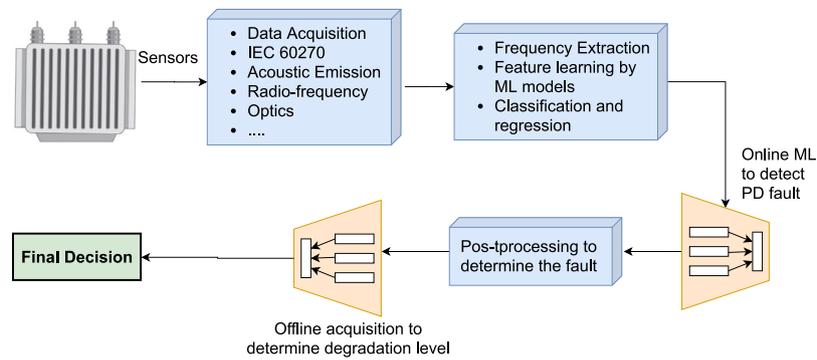

**Fig. 3.** Online or offline condition monitoring based on partial discharge activity using machine learning (Karthikeyan et al., 2008).

A very appealing feature of ML, especially regarding the field of PD diagnosis, is its ability to diagnose and identify both online and offline faults. This reduces the reliance on human experts for interpreting faults, resulting in cost and labor savings. Offline training of ML with diverse fault data seems feasible, and once the ML algorithm is trained with fault data, it can rapidly identify and track actual amount of insulation deterioration, indicating sudden fault correction (Karthikeyan et al., 2008; Majidi et al., 2015).

Fig. 3 illustrates the structural overview of PD-based CM system with post-processing feature using ML. ML is applied in two phases: Online ML based system for identification of actual problem and Offline ML system/algorithm for evaluation of the actual rate of deterioration in the insulation. Both phases show promising degree of potential for enhancement of CM functions. The offline PD degradation evaluation involves frequently trained and testing of ML algorithm, which contributes to lowering detection time and boosting diagnostic accuracy. Consequently, the integration of ML in PD-based monitoring systems can significantly impact maintenance costs and improve the overall dependability of power transformers (Karthikeyan et al., 2008; Majidi et al., 2015).

### 2.3. Breakdown voltage test

The BDV test is a crucial measure of the insulation strength in power transformers (Sai et al., 2020). It determines the voltage at which the transformer oil sample becomes conductive, indicated by a spark, and is typically reported in [kV/mm] units. BDV serves as an indicator of the insulation's ability to withstand electrical stress (Hadjadj et al., 2013).

During the BDV test, transformer oil is located in a test cell consisting of hemispherical electrodes, following the guidelines provided by IEC 60156 and IS 6792 (Sai et al., 2020). The test involves the application of an electric field to the oil sample. In the initial phase, thermal aging occurs in oil, which will lead to development of microscopic cavities. With further electric field application, gas development and microbubble formation take place. Continual exposure to electric field results in gas production rate which exceeds bubble formation, leading to development of disruptive discharge. That specific voltage at which this disruptive discharge occur is known as breakdown voltage (IEC, 2010).

A widely used instrument for characterizing oil samples in the BDV test is the BA75 analyzer. A breakdown voltage tester in given in Fig. 4. The BDV value is negatively correlated with the aging of transformer oil. However, it is important to note that low BDV value does not primarily imply aging, but it can indicate the severe levels of impurity in the transformer oil, following BS 60422 guidelines (EN, 2013). Disruptive discharge, which occurs during the BDV test, indicates an insulation failure in the test instrument. This discharge completely shortens the insulation under test, thereby reduction in voltage between two test electrodes to zero. With respect to IEC 60060-1 standard, this discharge sometimes can also occur momentarily and commonly referred as non-sustained disruptive discharge (IEC, 2010).

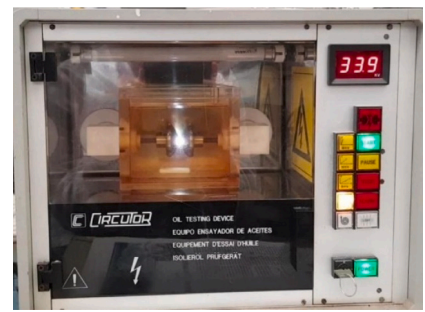

**Fig. 4.** Breakdown voltage tester (Monzón-Verona et al., 2021).

Different types of transformers have different minimum acceptable BDV values. For transformers operating at 230 kV, minimum acceptable BDV rate is 30 kV/mm. For transformers rated between 69 KV and 230 KV, the minimum value is 28 kV/mm. For transformers rated at or below 69 KV, the minimum acceptable BDV value is 23 kV/mm (Hayber et al., 2021). A Scoring Weighing for Oil quality including BDV, interfacial tension, moisture content, Acidity and color based on the calculation by TAM field experts is given in Table 4.

### 2.4. Photoluminescence (PL), Fourier-transform Infrared spectroscopy (FTIR) and Ultraviolet-Visible Spectroscopy (UV–VIS)

Spectroscopy is a discipline focused on measurement and investigation of spectra, which are plots of light intensity measured against specific properties of light, such as wavelength or wavenumber (Smith, 2011). It involves the study of matter's interaction with or emission of electromagnetic radiation. In particular, Infrared (IR) spectroscopy is concerned with the interaction between IR radiation and matter, based on the phenomenon of absorbance. This technique is valuable because different types of chemical structures produce different spectral fingerprints in the IR region. Fourier-transform Infrared spectroscopy (FTIR) is commonly used for such analyses due to its ability to identify functional bonding groups in molecules shown up in oil samples (Sai et al., 2020; Smith, 2011).

In the context of transformer oil analysis, FTIR plays a crucial role in characterizing the oil's composition. As oil ages, various chemical bonds undergo oxidation and thermal decomposition, leading to the formation of acidic and peroxide contents (Foo and Ghosh, 2002). FTIR analysis allows the identification and quantification of specific functional groups related to these chemical changes. Some of the functional groups analyzed include O-H (hydroxyl groups comprising carboxylic acids and alcohol), C=O (carbon monoxide or carbonyl group) and C-H (methine group).

Table 4
Scoring weighting for oil quality (Rediansyah et al., 2021).

| Oil quality | Score | | | | Weight |
|---|---|---|---|---|---|
| | 1 | 2 | 3 | 4 | |
| Breakdown Voltage (KV) | > 50 | 50–45 | 45–40 | <40 | 0.169 |
| Water Content (PPM) | <20 | 20–25 | 25–30 | > 30 | 0.108 |
| Acidity (mgKOH/mg) | <0.1 | 0.1–0.15 | 0.15–0.2 | >0.2 | 0.139 |
| Interfacial Tension (Dyne/cm) | > 35 | 35–25 | 25–20 | <20 | 0.124 |
| Color Scale | <1.5 | 1.5–2 | 2–2.5 | >2.5 | 0.114 |

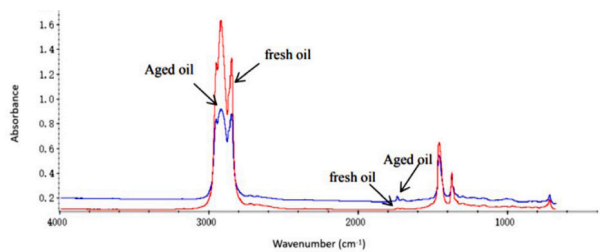

**Fig. 5.** FTIR spectral analysis for aged and fresh transformer oil (Liu et al., 2019).

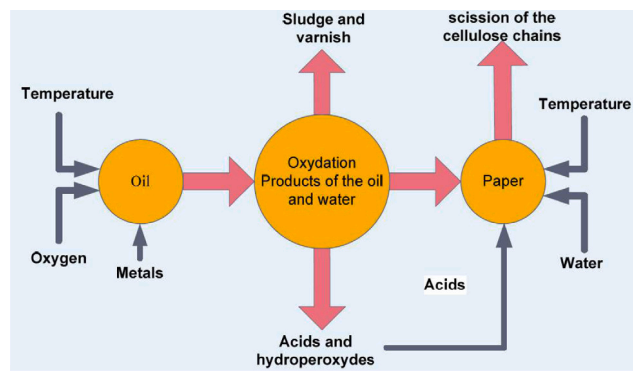

**Fig. 6.** Simplified decomposition mechanism of transformer insulation (Hadjadj et al., 2015).

The interpretation of FTIR spectra is facilitated by an IR interpretation table, which helps recognize functional groups based on the specific frequency at which atoms in the oil vibrate. When IR frequency matches an atom's frequency, a peak is formed in the spectrum, allowing for the identification of particular functional groups, reported in Sai et al. (2020) and shown in Fig. 5. In aging transformer oil, the intensity of peak absorbance for certain functional groups changes. According to findings of an FTIR analysis carried out by Alshehawy et al. (2017), the intensity of peak absorbance for the methane and carbonyl groups tends to increase with aging, while the intensity of the hydroxyl group peak decreases with aging.

UV–Vis and PL spectroscopy are non-intrusive and non-destructive methods used to probe materials (Alshehawy et al., 2021a). PL spectroscopy involves measuring the intensity of energy of light emitted while electronic transitions from some excited to ground state. This technique allows for the measurement of optical fluorescence as a function of wavelength, and it has the advantage of being straightforward and sensitive, particularly due to its narrow band of electronic states. PL spectroscopy can also be implemented online using available PL sensors (Alshehawy et al., 2021b). The configuration for PL spectroscopy typically includes a laser source, a cuvette to hold the oil sample, an optical lens to converge the laser light, a monochromator to select narrow bands of light wavelength, a detector acting as an amplifier, a passive transducer and a workstation for analysis of signal (Alshehawy et al., 2021a).

On the other hand, UV–Vis spectroscopy is extensively used for transformer oil condition assessment, but it has faced some criticism due to its cost and limited sensitivity to fluorescence (Alshehawy et al., 2021b). UV–Vis spectroscopy measures optical absorption as a function of wavelength. However, recent studies have shown that PL spectroscopy has demonstrated better correlation with the dissolved gas analysis-derived degree of polymerization compared to UV–Vis spectroscopy (Alshehawy et al., 2021a).

Both PL spectroscopy and UV–Vis spectroscopy play significant roles in transformer oil analysis and provide valuable information on the condition and aging of the oil (Alshehawy et al., 2021a). FTIR spectroscopy proves to be a valuable tool for offline transformer oil characterization.

By analyzing the infrared spectra of transformer oil samples, it provides critical insights into the chemical composition, aging processes, and degradation of the oil. This information is essential for assessing the health and condition of power transformers and making informed decisions regarding maintenance and replacement strategies.

### 2.5. Total acid number

The TAN is a crucial indicator of the acid concentration in transformer oil insulation, and it is strongly associated with the aging process (Hadjadj et al., 2013; IEC, 2010; IEEE, 2016). A decomposition mechanism of transformer is shown in Fig. 6 It is also known as the Neutralization Number (NN) and is determined by measuring the amount of potassium hydroxide (KOH) required to neutralize the acid present in one gram of a transformer oil sample. The TAN value is expressed in milligrams of KOH per gram of sample (mgKOH/g) (Hadjadj et al., 2015). According to the standard BS 62021-1, the TAN represents the quantity of base (KOH) needed to potentiometrically titrate a specific test portion of the oil sample dissolved in a solvent to obtain a pH of 11.5 (IEC, 2003).

The TAN measurement is essential for monitoring the condition of transformer oil, as oil with a TAN value less than the marginal class of Oil Quality Index Number (OQIN) reference is generally considered unsafe for continual usage and it should be reclaimed (Hadjadj et al., 2015). To determine the TAN, standards such as BS 62021-1 and BS 62021-2 specify various methods for mineral transformer oil acidity determination. BS 62021-1 outlines the procedure for automatic potentiometric titration, while BS 62021-2 deals with colorimetric titration (IEC, 2003; BSI, 2007). In the automatic potentiometric titration method described by BS 62021-1, the sample of transformer oil is dissolved in some solvent and then titrated by using a glass-indicating electrode and some reference electrode. Unlike other methods, potentiometric titration does not primarily require an indicator; infact, it involves measuring the potential to determine the endpoint of the titration, which is specified as reaching a pH of 11.5. Volume of the base consumed during the titration to reach this endpoint is reported as the NN (IEC, 2003).

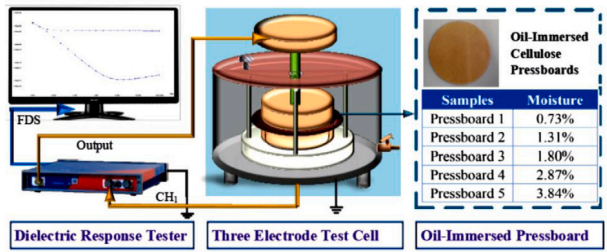

Fig. 7. Frequency domain spectroscopy (Liu et al., 2020).

### 2.6. Frequency domain spectroscopy and frequency response analysis

When some transformer experiences high interrupting fault currents, it undergoes significant mechanical forces that can lead to movement and deformations in its mechanical structure and windings. Traditionally, detecting winding deformations in transformers has been challenging using conventional diagnostic techniques. However, the FRA method has emerged as an advanced diagnostic approach that can detect such winding faults with sensitivity and repeatability.

In FRA, the phase margin and amplitude of winding impedance, transfer function and admittance are plotted against frequency. This allows for the observation of relatively minor changes in winding capacitance and inductance caused by deformations (Yin et al., 2020). To detect winding damage using FRA, the measured FRA trace is compared to its fingerprint corresponding to healthy state of transformer winding or estimated response by use of an equivalent circuit model (Yin et al., 2020).

Similarly, FDS is an advanced diagnostic technique used in the evaluation of power transformers. It involves the analysis of the dielectric response of the transformer insulation over a wide range of frequencies. FDS is a powerful tool for detecting and evaluating various conditions in transformer insulation, including moisture content, aging, and faults (Liu et al., 2020). The FDS measurement is based on the principle that the dielectric properties of the insulation vary with frequency, and these variations carry valuable information about the condition of the transformer. By subjecting the transformer insulation to an AC signal at different frequencies, the dielectric response is recorded and analyzed (Liu et al., 2020). The FDS method can cover a broad frequency range, from extremely low frequencies (mHz) to radio frequencies (MHz). Fig. 7 shows a schematic diagram for FDS.

The challenge with FRA and FDS interpretation lies in the need for expertise in the field, as there is no widely-accepted automatic interpretation algorithm (Samimi et al., 2016). Interpreting FRA/FDS data can be complex since obtained transfer function changes highly from one case to another, and different types of faults may have varying effects on the transfer function. However, the application of ML algorithms has shown promise in automatically analyzing FRA and FDS data not only to recognize the presence of faults but also to determine the type and location of the fault (Picher et al., 2020). To develop an ML-based system for FRA interpretation, data from various failure scenarios emerging from events like experiments or models should be input in the system. System can then have trained on this data to find out subsequent faults in that FRA data accurately. ML-based FRA interpretation claims an accuracy of over 98% (Picher et al., 2020). This approach holds potential in providing an efficient and accurate means of detecting and diagnosing winding damage in transformers during the offline maintenance process, thereby preventing sudden failures and reducing losses.

### 2.7. Furan tests

The power transformer undergoes both thermal and electrical stress during its service life, leading to oil degradation and a reduction in some of its strength properties. To ensure the proper functioning of transformers and to detect any potential issues during service, various physical, chemical, and electrical tests are regularly conducted. These tests help make a database system which aids in knowing the deteriorating behavior the transformers experience and making informed decisions (Soni et al., 2021).

As transformers age or operate abnormally, the cellulose paper insulation also undergoes decomposition. For instance, overloading followed by high temperatures can lead to the development of furanic compounds in the cellulosic insulation, particularly when oxygen and moisture are present at high operating temperatures. The concentration of these furanic chemicals serves as a useful tool for prediction of aging and potential defects in the insulation. Several studies have explored the purpose of furanic compounds in understanding the aging process of cellulose paper in transformers, with one such study using High-Performance Liquid Chromatography (HPLC) being applied for aging analysis (Saha, 2003). The usual limits incorporated by ASTM and IEEE are provided in BSI (2007).ML techniques offer valuable insights, especially in transformer insulation quality evaluation, where data might be scarce or difficult to obtain. Degree of polymerization (DP) is a critical parameter to determine mechanical life for cellulose in transformers. When the DP drops below a certain threshold (typically 150–200), the mechanical strength of the paper reduces, and the insulation might fail in the event of a short circuit. Traditionally, quantifying DP required taking a paper sample from inside of transformer by draining transformer oil, which can be crucial in some cases and cause damage to transformer. Additionally, DP is highly critical in areas having higher temperatures, e.g. around the winding's top position, whereas sampling using a conductor paper is impossible because of the multi-layer insulation of winding. In such cases, estimating DP using easily measurable parameters becomes a viable option. Hence, some methods use furanic chemicals to estimate the DP number, as there are some relations between DP value and furfural/furanic compounds present in transformer oil paper insulation. But sometimes they may not provide a strong correlation between furanic chemicals and DP due to various other influencing factors present in DP value (Brochure, 2012; Hohlein and Kachler, 2005).

As Furanic compounds plays crucial role in determining the remaining life of transformers. However, measuring furanic compounds is often challenging and prone to various errors. Several parameters, such as water content, oil breakdown voltage, total combustible gases and acidity, among others, can influence the measurement of furanic compounds. To address this, researchers have explored the application of ML to determine the quantity of furanic compounds on the basis of other transformer oil quality parameters (Ghunem et al., 2012).

In a study described in Ghunem et al. (2012), a ML-based prediction model was developed to find out the total furan content in transformer oil using inputs from dissolved gases and oil quality parameters. The researchers tested this model on in-service power transformers and found that it achieved an impressive 90% accuracy in prediction of furan content present in transformer oil. This highlights the potential of ML-based models in significantly improving the reliability of estimating the remaining life of power transformers.

### 2.8. Fiber optic sensors

The thermal conditions inside a transformer and its ability to dissipate heat are crucial factors affecting the insulation's degradation and, ultimately, the transformer's operational longevity. To assess the thermal performance of power transformers, the strategic placement of optical fiber sensors for temperature monitoring is essential. An approach in Rodrigues et al. (2023) combining experimental and numerical methodologies is used. For instance, during the temperature rise test of a single-phase prototype transformer with an Oil-directed and Air-natural (ODAN) cooling system, 20 optical fiber sensors are deployed along the windings to gather temperature data.

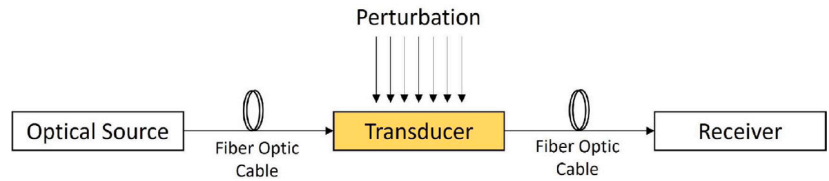

**Fig. 8.** A typical Fibre Optic Sensing system (Ramnarine et al., 2023).

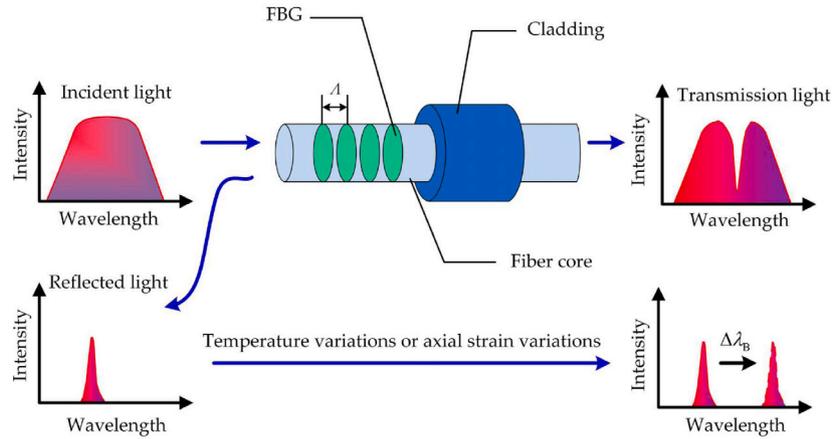

**Fig. 9.** Principle of Operation of FBG Sensors (Ramnarine et al., 2023).

Optical fiber sensors (OFS) are highly regarded for their rapid and precise fault detection capabilities in electrical equipment, ensuring consistent and stable operation. The block diagram of a typical optical fibre sensing system is shown in Fig. 8. Recent advancements have introduced a variety of optical fiber sensors, such as interferometric sensors, distributed sensors, spectroscopy sensors, and Fiber Bragg Grating (FBG) sensors (Ramnarine et al., 2023).

FBG sensors, which operate using single-mode fibers with periodic variations in the core's refractive index (known as gratings), have become particularly prominent in monitoring electrical equipment (Sun and Ma, 2023). Their ability to multiplex and cover a broad sensing range makes them ideal for high-voltage applications. Despite their advantages, the integration of FBG sensors in transformers is still advancing, offering potential improvements in measurement accuracy and overall equipment monitoring. Principle of operation of Fibre Bragg Grating Sensor is given in Fig. 9.

OFS provides precise temperature and strain measurements but typically require additional analytical methods to interpret data for fault diagnosis. Also OFS can detect temperature anomalies, they lack the sophisticated classification capabilities of ML models (Dureck et al., 2023) and typically need manual interpretation. ML methods excel in data analysis, adaptive learning, and handling complex, non-linear relationships, making them superior for enhancing transformer monitoring and maintenance strategies. ML algorithms can handle large datasets, identify complex patterns, and offer precise fault classification, which enhances the accuracy and reliability of fault detection. ML models also integrate well with Internet of things (IoT) and smart grid systems, providing real-time data processing and scalability for comprehensive monitoring.

## 3. Power transformer remaining life assessment

Assessment of remaining transformers life span is crucial task in transformer diagnostics, as transformers undergo aging due to various factors. The aging of transformers largely depends on the insulation system, making the quality of the insulation paper a key factor in determining their remaining lifespan. Several causes contribute to the aging process. Firstly, temperature owns a significant role in the degradation of insulating materials, specifically cellulose. Secondly, moisture disturbs the cellulose chain and leads to various chemical reactions. Thirdly, the presence of oxygen mostly causes oil to oxidize, triggering many more degradation processes. Fourthly, mechanical stresses weaken the strength of oil paper and pressboard components. Lastly, the involvement of acids and other corrosive elements in the deterioration process further complicates the aging of transformers. A brief Transformer oil classification considering 4 factors is given in Table 5 (where OQIN stands for the oil quality index number). These factors are interconnected and, in some cases, not fully understood, making it challenging to accurately characterize the remaining life of transformers (Forouhari and Abu-Siada, 2018).

The health of a transformer is primarily determined by the condition of its solid insulation, which consists of a linear cellulose polymer. The DP refers to the number of monomer units in this polymer. In new transformers, the DP value of the solid insulation ranges from 1100 to 1600. However, thermal, hydrolytic, and pyrolytic stresses can degrade the polymer, reducing the DP value. A DP value of 200 typically signifies the end of the insulation's useful life. Aging by-products such as water, furan, carbon dioxide, carbon monoxide, and acids are produced during this process, further accelerating insulation aging (Li et al., 2018). The aging process of transformer solid insulation (cellulose polymer) is shown in Fig. 10. The relationship among Remaining Useful Life (RUL), Loss-of-Life (LOL), and Low Temperature Over Temperature (LTOT) of a transformer is given in Fig. 11.

In Hillary et al. (2017) DP is presented as a critical parameter that indicates the condition of the insulation paper, typically measured using furan analysis. Various relationships between DP and furan content are utilized in the power industry to calculate a transformer's remaining lifespan. A mathematical model was developed in this paper using multiple linear and nonlinear regression techniques to estimate the furan content of a transformer, and thereby its remaining lifespan. Key factors influencing transformer aging, such as moisture content, temperature, transformer capacity, oxygen content, breakdown voltage, and current age, were identified and incorporated into the model. The mathematical model establishes a correlation between these factors and furan content. The DP value is then derived from the estimated furan

**Table 5**
Transformer oil classification (Hadjadj et al., 2015).

| Sr.No. | Oil quality | AN | IFT | Color | OQIN |
|---|---|---|---|---|---|
| 1 | Good Oils | 0.00–0.10 | 30.0–45.0 | Pale Yellow | 300–1500 |
| 2 | Proposition A Oils | 0.05–0.10 | 27.1–29.9 | Yellow | 271–600 |
| 3 | Marginal Oils | 0.11–0.15 | 24.0–27.0 | Bright Yellow | 160–318 |
| 4 | Bad Oils | 0.16–0.40 | 18.0–23.9 | Amber | 45–159 |
| 5 | Very Bad Oils | 0.41–0.65 | 14.0–17.9 | Brown | 22–44 |
| 6 | Extremely Bad Oils | 0.66–1.50 | 9.0–13.9 | Dark Brown | 6–21 |
| 7 | Oils in Disastrous condition | 1.51 or more | – | Black | – |

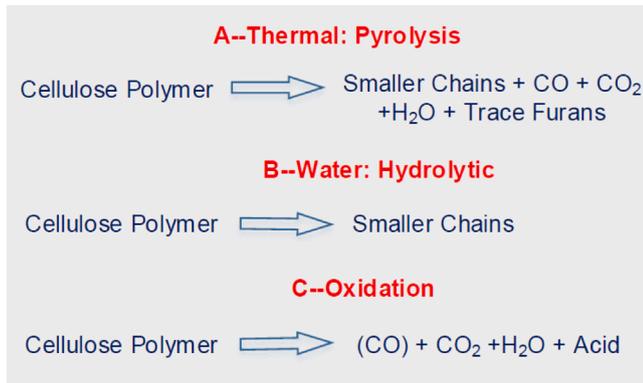

**Fig. 10.** Transformer solid insulation aging process (Li et al., 2018).

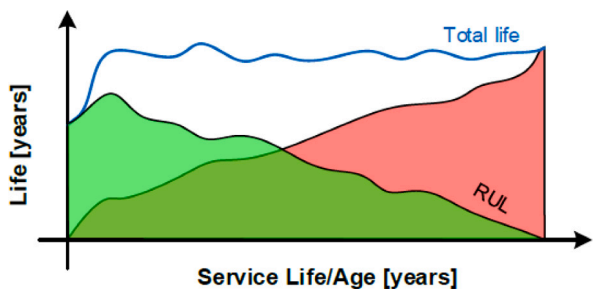

**Fig. 11.** Transformer remaining useful life illustration (Li et al., 2018).

content, allowing for the calculation of the transformer's remaining lifespan.

Load and ambient temperatures are critical factors affecting transformer insulation lifespan. Muthanna et al. discusses hourly monitoring data to estimate load factors and ambient temperatures based on historical data. These estimates are applied to IEEE life consumption models to evaluate the insulation's consumed life. Two methods, full simulation and renewal process approaches, are proposed to predict insulation failure time and estimate reliability parameters such as time to reach design life, and failure probability. They emphasizes that the full simulation approach, while computationally intensive, accounts for long-term variations, whereas the renewal approach simplifies the process but needs to include long-term shifts in conditions for accurate reliability assessment (Muthanna et al., 2005).

In Foros and Istad (2020), a structured approach for evaluating and ranking the condition of power transformers is described. It allows for the comparison of transformers, identification of those needing attention, and provides recommendations for maintenance or replacement. The method integrates three fundamental models: a physical winding degradation model, a health index model that leverages condition monitoring data and expert judgment, and a statistics-based end-of-life model. The statistics-based model utilizes data from a database of decommissioned transformers being developed in Norway. By merging the first two models with the statistics-based model, it calculates an individualized and condition-dependent probability of failure.

This approach allows for the estimation of the expected remaining lifespan. In Pandurangaiah et al. (2008) the author focus on Data-driven Diagnostic Testing and Condition Monitoring (DTCM) for power transformers. The authors have developed a Monte Carlo approach to supplement the limited experimental data typically gathered from prototype transformers. They also describe a validation procedure to assess the accuracy of the developed model. The importance of data acquisition, which is critical in DTCM for power transformers, and the generation of data using Monte Carlo techniques are highlighted. The generated data's validity is confirmed through statistical significance tests. Additionally, an empirical regression model is created to estimate both the elapsed life and the remaining life of power transformers.

ML techniques offer a promising solution to account for all these aging effects. An expert system, as shown in Fig. 12, utilizes available data from transformers to estimate their remaining life (Bakar and Abu-Siada, 2016).

In the first phase of the model, the system is trained using data from those transformers which have reached the end of their healthy life span. Second step involves, extraction of data from existing transformers and is fed into the system to calculate the remaining life. Moreover, if important data is unavailable, then trained system can make predictions using alternative data. One proposed ML-based approach uses insulation oil tests to predict the remaining operational life of power transformers. This intelligent model relies on field data collected from various transformers and has been shown to be reliable, providing timely asset management decisions with reduced dependence on expert personnel (Bakar and Abu-Siada, 2016).

Hence, predicting the remaining life of transformers is a complex task due to the interconnected nature of various aging factors. A review of current methods for diagnosing transformer issues, highlighting the drawbacks and limitations of traditional approaches is presented in Zou et al. (2022). Traditional methods rely on static data and fail to provide real-time mapping to objects, leading to potential delays in detection and significant errors. To address these challenges, data-driven methods for transformer fault diagnosis will be very useful. In this context, ML techniques offer a promising approach to consider these effects and develop expert systems that can accurately estimate the remaining life of transformers based on available data, enhancing asset management decisions in the field of transformer diagnostics (Forouhari and Abu-Siada, 2018; Bakar and Abu-Siada, 2016).

## 4. Machine learning models

ML models, specifically regression and classification models, play a crucial role in obtaining online ageing prediction for transformer oil or paper insulation. Regression models are employed to predict key ageing properties like TAN, IFT, Decayed Dissolved Particles (DDP), turbidity, DDF, and more, based on the sensor's output variable. On the other hand, classification models are used to differentiate various levels of severity by utilizing the OQIN value (Elele et al., 2022). These models enable prediction based maintenance options by alerting operators of thresholds reaching dangerous levels and providing initial online recommendations for maintenance actions.

Several ML regression models are suitable for online ageing detection, including nonlinear regression, linear regression, decision trees regression, Support Vector Machine (SVM) regression, and shallow/deep neural network regression models. These models map the target variable(s) (e.g., IFT, TAN, DDF) to one or more predictor variables, typically obtained from the sensor's output data. On the other hand, ML classification models, such as decision trees, logistic regression, SVM, K-Nearest Neighbour (KNN), and shallow/deep neural networks, are used to classify the severity levels of ageing into different classes or categories based on the OQIN value (Elele et al., 2022).

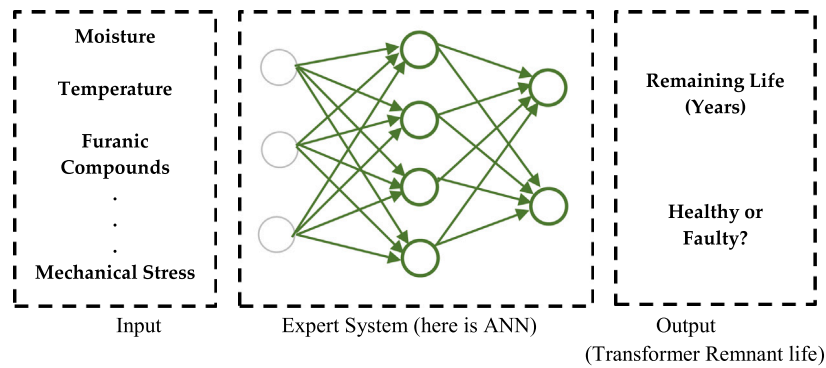

**Fig. 12.** Determination of transformer remnant life predicted using machine learning (Ghunem et al., 2012).

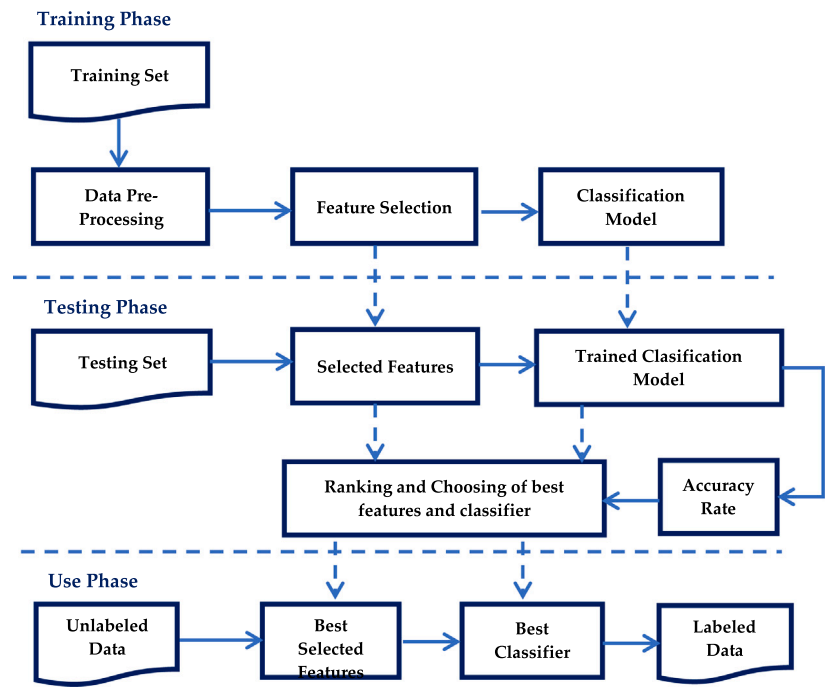

**Fig. 13.** The process flow of developing ML based models (Bhatia et al., 2020).

The regression model is particularly familiar, mapping the target variable(s) to predictor variable(s) from the use of linear functions, as seen in simple linear regression (one predictor variable) and multiple linear regression (multiple predictor variables). These regression models operate under certain assumptions, including linearity, independence, homoscedasticity, and normality in between predictor and target variables. The process of developing ML based models is given in Fig. 13.

Non-binary classification and regression models can be created using neural networks and decision trees. Decision trees are graphical structures resembling trees, where each leaf node represents the outcome of a series of decisions, and each branch node signifies a choice among multiple options (Tong and Ranganathan, 2013). Artificial Neural Networks (ANN) are a type of supervised ML techniques that imitate biological networks, using interconnected layers to process data. For a detailed exploration of artificial neural networks and decision trees, refer to Cheng and Titterington (1994).

ML models, when combined with sensor data and the IoT, offer a prominent maintenance solution for transformer oil ageing. IoT data collection often generates sparse, inconsistent, and noisy data, which can reduce the confidence and certainty in data analysis. To improve the reliability of data analysis, it is crucial to manage and quantify model uncertainty. This approach not only enhances the confidence in the model's judgments but also strengthens its overall dependability (Polužanski et al., 2022). In-depth studies on uncertainty quantification for deep learning and ML models reveal the utilization of Bayesian physics informed networks and Bayesian neural networks for deep learning uncertainty quantification. Additionally, physics-informed neural networks and Gaussian process regression (GPR) are employed for traditional ML (Siddique et al., 2022). By analyzing the sensor data in real-time and applying ML algorithms, operators can make informed decisions about maintenance actions, preventing catastrophic failures and optimizing transformer performance. The basic principle of AI based algorithms for predicting Health Index in shown in Fig. 14.

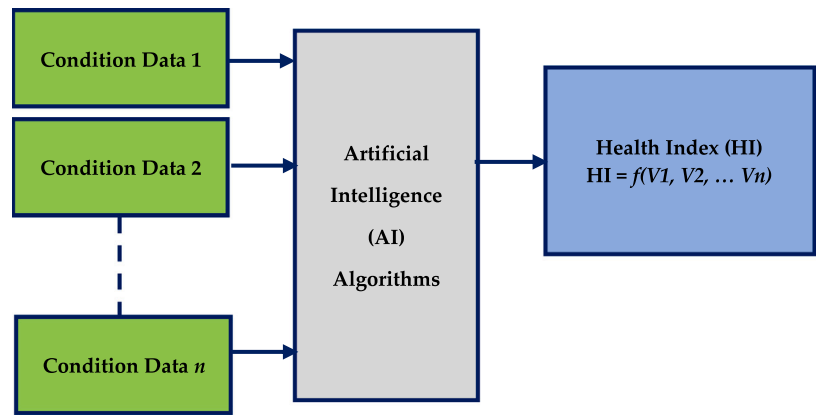

**Fig. 14.** Principles of AI-based transformer health index prediction (Rediansyah et al., 2021).

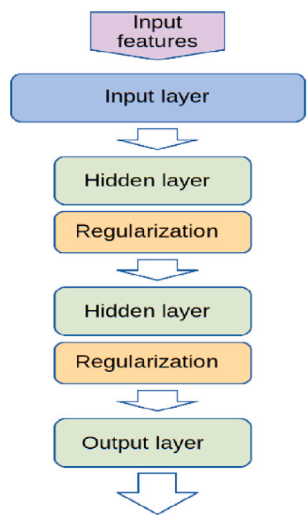

**Fig. 15.** Bayesian feed-forward artificial neural network (Sarajcev et al., 2019).

*4.1. Deep learning models*

Deep learning is significantly advanced machine learning towards artificial intelligence. Unlike traditional "shallow learning" methods, deep learning involves multiple layers of nonlinear operations. The features learned by deep learning models are more representative of the original data, greatly aiding in classification and visualization tasks. A block diagram of Bayesi an artificial neural network architecture showing various layers is given in Fig. 15. Consequently, deep learning has become increasingly popular among researchers in transformer fault diagnosis. Although much work has not been conducted on deep learning models in terms of transformer health index and life time estimation but a few researchers provide valuable contribution in this context (Zhang et al., 2022).

In Islam et al. (2023) a machine learning framework aided by deep generative models to evaluate the health of high-voltage power transformers is presented. Using a dataset of 31 input parameters from 608 transformers, various machine learning models were initially applied but showed low accuracy due to high dimensionality and limited data. To address this, different types of Autoencoder (AE) were used for feature extraction and dimension reduction, including single-layer AE, sparse AE, Multi-Layer Perception (MLP) AE, stacked AE, and stacked-sparse AE. After compressing the data with the AE, the models were re-evaluated, and it was found that the combination of MLP AE and various classifiers significantly improved accuracy, with logistic regression paired with MLP AE achieving the best results. This proposed model outperformed existing models in accuracy.

A Bayesian "Wide and Deep" machine learning model is introduced in Sarajcev et al. (2019) for HI calculation using transformer data. This technique combines Bayesian ordered robust regression (wide component) and a Bayesian artificial neural network (deep component). Both parts are trained together using a Markov-chain Monte Carlo algorithm. Unlike traditional regression models that often produce out-of-range HI values, this model categorizes HI using expert-defined, ordered categories. It demonstrates equal or superior accuracy compared to previous models and fully quantifies parameter and prediction uncertainties. The model is validated with real transformer data. Additionally, the model can be adapted for datasets with distribution and transmission transformers, incorporating a hierarchical structure for parameter sharing across groups. It also supports sequential (online) learning, making it ideal for continuous transformer health monitoring. These features enhance its suitability for transformer health analysis (Sarajcev et al., 2019). Another paper addresses the challenges of limited fault sample data and data imbalance by proposing a new data augmentation method using Kernel Principal Component Analysis (KPCA). This technique maps original data into a high-dimensional feature space to create new, similar samples. Additionally, a deep residual network with an identity path is introduced for fault diagnosis, improving the transfer and update of weight parameters. Simulation results show that the method effectively expands data samples and enhances fault diagnosis accuracy through strong feature extraction capabilities (Liu et al., 2023).

*4.2. Machine learning techniques*

In Fig. 16, an overview of the main ML techniques and features is presented. Supervised machine-learning algorithms aims for modeling correlations and dependencies between prediction outputs and input properties. They utilize prior data sets to determine correct outputs for new data based on learned relationships. Three common applications of supervised learning are prediction, regression, and classification. One prominent example of supervised learning is ANNs, extensively employed for detecting and categorizing defective states (Chawla et al., 2005).

On the other hand, unsupervised learning allows machines to explore data without prior labels. It seeks hidden patterns connecting distinct variables after an initial investigation. Unsupervised learning aids in data grouping using simple statistical features (Li et al., 2017). Unlike supervised learning, it does not require large data sets for training, hence, making it easier and faster to implement algorithm. Moreover, unsupervised learning can indicate inaccurate data which does not fit into proposed categories.

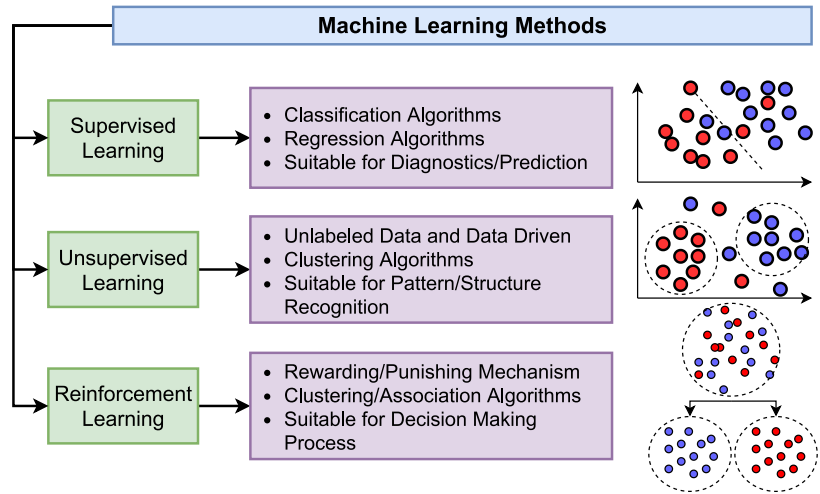

**Fig. 16.** Overview of machine learning techniques (Esmaeili Nezhad and Samimi, 2022).

Reinforcement learning is Data training technique which improves desirable functions while deleting undesirable ones. An agent of reinforcement learning interprets its learning environment by taking actions and learning via trial and error. Subsequent paragraphs introduce most widely applied ML algorithms used for transformer condition assessment and life span prediction in recent studies, along with their advantages and disadvantages specifically in field of transformer health analysis and diagnosis.

*4.3. Artificial neural network*

The ANN is one of the most widely used ML algorithm for fault diagnosis of transformer (Leal et al., 2009; Farag et al., 2001). It emulates the structure of brain neurons just like humans, offering significant information processing parallel abilities, fault tolerance, resilience, and self-learning abilities. The ANN can effectively map highly unknown and nonlinear input–output correlations in systems. The back-propagation approach is extensively used in supervised learning to train feed-forward neural networks, resulting in what is known as Back-Propagation Neural Network (BPNN). Generally, BPNN is one of most widely used branch of ANN for diagnostic applications.

Numerous work in the literature have employed ANN to address transformer failure diagnostic problems. For instance, in Abu-Elanien et al. (2011), an ANN method is utilized to determine the transformer's condition based on predicted HI (Hydrogen to Nitrogen) value. In a model, a feed-forward ANN having 2 hidden layers (4 and 2 neurons, respectively), is trained using actual data from 59 transformers. Results show that 97% of the total test samples are accurately identified on the basis of three-class condition. Another work presents an enhanced transformer fault diagnosis method using a residual BPNN. This approach deepens the BPNN by incorporating multiple residual network modules. It improves and extends the Dissolved gas feature information analysis through the enhanced BPNN. This method employs a multi-layer neural network to extract more distinct gas feature information after fusion, significantly enhancing diagnostic accuracy. Experimental results indicate that this proposed algorithm outperforms traditional BPNN methods, achieving a diagnostic accuracy rate of 92%, thus ensuring the continuous, stable, and safe operation of power grids (Jin et al., 2023).

Similarly, other research, including (Leal et al., 2009; Farag et al., 2001), has also employed the same ANN method for transformer failure diagnostics. In Zhang et al. (1996), Zhang et al. propose an ANN system on the basis of DGA for the recognition and determination of faults in power transformers. They utilize a two-step ANN framework, where the first type of ANN determines the nature and location of fault, and the second type determines whether or not any contribution from cellulose. The accuracy for fault diagnosis of ANN in this research is shown to be well-performing.

*4.4. Support vector machine*

SVM is an influential set for supervised learning method used in purpose of classification, detection and regression tasks. This algorithm excels in handling both linear and nonlinear applications and performs well even with limited amounts of data. SVM works by finding the best planes that can disseminates a dataset into a number of groups or approximate single function (Gholami and Fakhari, 2017). The algorithm creates areas on the basis of existing vectors in order to classify data and uses this classification to analyze new data. One of the key advantages of SVM is its robustness in handling classification and regression issues compared to other ML approaches like ANN. Unlike some other algorithms, SVM always seeks some global solution instead of some local one. However, one of the challenges with SVM lies in the selection of suitable parameter values, as classification process heavily depends on these parameters to achieve optimal results.

Support Vector Regression (SVR) is a sub branch of SVM used for predicting numerical property values, for example compound potency. As a substitute of constructing a hyperplane used for labeling class prediction, SVR derives a different function based on training data for prediction of numerical values. Similar to SVM, use of kernel functions is also done by SVR to map data into higher-dimensional feature spaces, allowing it to handle nonlinear Structure-Activity Relationship (SAR) effectively. This characteristic makes SVR particularly appealing for potency prediction, as it is not limited to the applicability domain of traditional quantitative SAR analysis methods (Aizpurua et al., 2018). However, one drawback of both SVM and SVR is their black box nature, meaning that their predictions cannot be easily converted into chemical terms.

In recent years, SVM has been increasingly combined with other algorithms to develop new and improved ML algorithms. For instance, in Siddique et al. (2022), Fei and Zhang proposed integrating SVM in cooperating a Genetic Algorithm (GA) for creation of SVM-GA method for diagnosing defects in power transformers. The experimental findings demonstrated that this innovative approach outperformed IEC three-ratio method, standard SVM classifier and ANN in terms of accurate diagnosis. Similarly, in Zheng et al. (2011), Zheng et al. proposed a two-classifier cascade method for power transformer fault determination, addressing single and multiple failure determinations. In described approach, SVM served as 1st classifier to categorize the power transformer's faulty or normal condition, while GA was used to enhance the SVM's kernel function parameter. The results showed that this combined approach allowed for the classification of power transformers having multiple or single with acceptable accuracy by use of genetic SVM method.

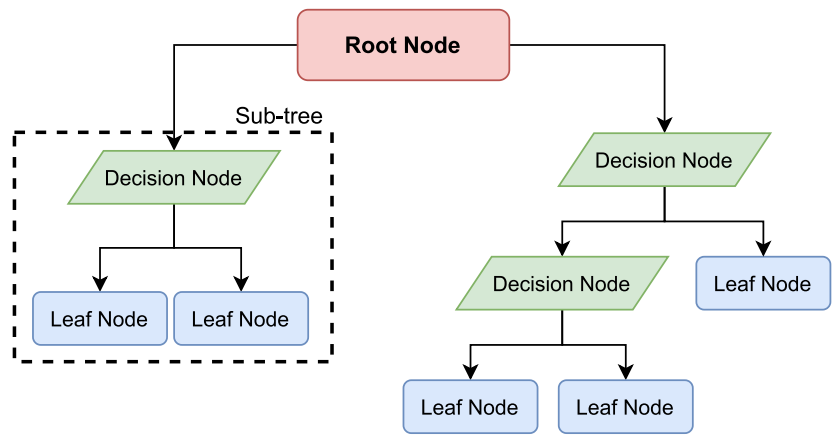

**Fig. 17.** A decision tree based structure for classification problem (Esmaeili Nezhad and Samimi, 2022).

*4.5. Random forest*

Random Forest (RF) is an advanced tree-based algorithm that enhances Decision Trees (DT) by combination of multiple DT-based classifiers and mitigating over-fitting as given in Fig. 17. RF achieves this by training numerous decision trees and predicting the class based on majority votes (Shil and Anderson, 2019). Due to its strong generalization ability, RF has been effectively utilized for fault diagnosis in transformers (Chen et al., 2011). The RF model can develop a proximity matrix on the basis of pattern similarity without requiring data preprocessing. Several tests on real transformers have shown that the RF-based approach performs much better than traditional classifiers like SVM in terms of diagnostic accuracy (Chen et al., 2011).

Moreover, in another research, RF was employed to compare and evaluate total 91 samples obtained from oil power transformers. Main goal was to check whether improvement taken place in decision accuracy compared to prior algorithms like artificial neural networks (ANN). Results indicated that RF performed better than ANN, irrespective of data size (Senoussaoui et al., 2021). The advantage of RF-based methods over previous algorithms like ANN and SVM lies in their ability to achieve superior diagnostic accuracy with less data, making them more suitable for transformer diagnostic programs (Senoussaoui et al., 2021).

*4.6. K-nearest neighbour*

The KNN algorithm is a straightforward and valuable data mining algorithm used for both classification and regression tasks. It works by classifying a new data point on the basis of class labels for its KNN in feature space. The KNN technique assumes similarity in between a new and old case and assigns new case to category which is most likely to match existing categories. This algorithm's outputs are class labels, and in regression applications, it predicts average value of K-nearest neighbors' outputs (Cover and Hart, 1967; Hussein et al., 2017).

KNN provides two prominent advantages. Firstly, it is a simple and easy-to-understand ML model, making it an attractive choice for beginners in the field of ML. Secondly, it is a non-parametric algorithm, meaning it does not assume any underlying data distribution, making it flexible and applicable to various scenarios. Additionally, KNN does not require any training process, making it suitable for real-time applications with continuously generated data. It can handle large datasets without suffering from the curse of dimensionality, making it a suitable choice for high-dimensional data problems. Furthermore, KNN is known for its accuracy and effectiveness, especially with small to medium-sized datasets, as it can handle noisy and incomplete data (Hussein et al., 2017).

*4.7. Genetic algorithm*

The GA is a powerful computer program used for optimizing and supporting multiple-objective functions. It has been widely applied in various cases to overcome optimization challenges. In Jaiswal et al. (2018), a hybrid approach for transformer HI calculation is proposed, combining the transformer Health Index obtained using weighted parameters with the genetic algorithm. GA is used to enhance the conventional Health Index calculations by optimizing these weighted specifications. Similarly, in Fei and Zhang (2009) and Kari et al. (2018), GA is combined with SVM to overcome weaknesses of an alone learning algorithms. Experimental results demonstrate that the combination of GA with SVM increases accuracy of diagnosis process in comparison to using a conventional SVM classifier (Jaiswal et al., 2018). The blend of GA and SVM has proven to be effective in addressing the challenges associated with fault diagnosis and feature selection for power transformers.

*4.8. Convolutional neural network*

A model proposed in Taha (2023) employs a Convolutional Neural Network (CNN) to predict and diagnose the HI of power transformers. An imbalance in the training dataset initially results in better predictions for the majority class and poorer detection for the minority class. To address this, an oversampling technique is used to balance the training data, thereby improving the accuracy of the classification methods. After applying oversampling, the CNN model predicts the HI of power transformers effectively. The performance of the proposed CNN model is then compared to optimized ML classification methods, with the CNN demonstrating superior results. Feature reduction techniques are implemented to decrease testing time, effort, and costs. Finally, the CNN model's robustness is evaluated with uncertain noise levels in both full and reduced feature sets, up to ±25%, showing reliable prediction of the power transformer HI. Classification model procedure using CNN model training is given in Fig. 18. A comparative table in terms of performance matrices between CNN and many other machine learning techniques is given in Table 6.

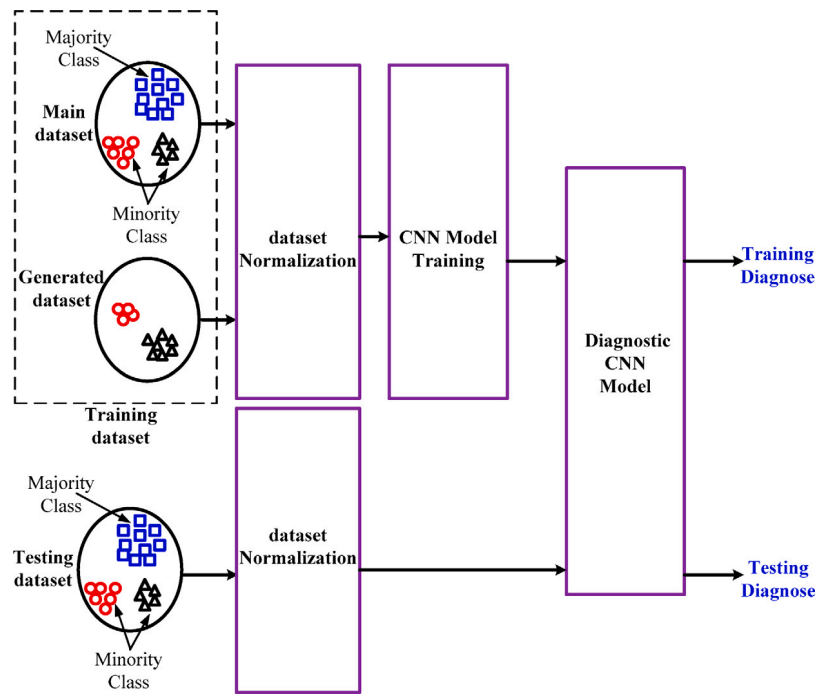

**Fig. 18.** Classification model procedure (Taha, 2023).

Table 6
Comparison between the results of the CNN model and other methods (Taha, 2023).

| HI | Good | Fair | Poor | Sensitivity | Specificity | Precision | F1-Score | % Accuracy |
|---|---|---|---|---|---|---|---|---|
| DT | 332 | 70 | 4 | 0.61 | 0.88 | 0.64 | 0.62 | 85.29 |
| SVM | 330 | 93 | 3 | 0.65 | 0.92 | 0.75 | 0.68 | 89.50 |
| KNN | 339 | 75 | 3 | 0.61 | 0.88 | 0.83 | 0.66 | 87.61 |
| ANN | 332 | 70 | 4 | 0.66 | 0.91 | 0.81 | 0.71 | 89.92 |
| CNN | 338 | 65 | 5 | 0.74 | 0.91 | 0.80 | 0.77 | 89.92 |

### 4.9. Particle swarm optimization

Particle Swarm Optimization (PSO) is a widely used optimization technique often combined with ANN to improve their performance. In scenarios where GA are employed, PSO can serve as an alternative optimization approach. Tang et al. (2008) developed a diagnostic classifier for transformer failures that utilizes a method to determine transformer DGA data on the basis of probability. PSO is applied for being global optimizer technique in research work for enhancing the accuracy for fault classification by optimizing classifier's parameters. Comparative analysis with various defect classification methods reveals that this strategy significantly enhances both computing efficiency and diagnosis accuracy. Additionally, in Chawla et al. (2005), PSO is combined with SVM to obtain optimal estimators for the SVM module, resulting in the generation of an improved classification model. The study demonstrates that this proposed method substantially increases the accuracy of transformer health evaluation. Notably, PSO-based techniques effectively eliminate redundant input parameters that may confuse classifier, thereby enhancing the system's efficiency. Similar approaches utilizing PSO can also be found in other works (Sarajcev et al., 2018; Illias et al., 2015).

### 4.10. Additional techniques

Various techniques are also applied in literature for different applications in transformer fault detection and prediction, particularly in field of PD determination. These methods include the Gaussian Mixture Model (GMM), tensor-based classifier, Blind Signal Separation (BSS), Multiple Linear Regression (MLR), Support Vector Data Description (SVDD), Sparse Representation Classifier (SRC), Bayesian Network (BN), and Rough Set Theory (RS) etc.

## 5. Author's opinion and future work

The assessment of power transformer health and the prediction of their lifespan have long relied on conventional methods. Techniques like DGA and insulation resistance tests provide critical information on the internal state of the transformer. Yet, these methods are often reactive rather than proactive, identifying issues only after they have become significant. The periodic nature of these assessments can lead to undetected degradation between inspections, potentially resulting in unexpected failures. While these methods have a proven track record, their inherent limitations necessitate the exploration of more advanced techniques.

The advent of ML has brought a transformative shift in power transformer health assessment. ML techniques can analyze vast amounts of data, identifying patterns and correlations that are not apparent through conventional methods, excel at processing and analyzing high-dimensional data, inherently data-driven, reducing the reliance on subjective interpretations. ML models have provide clear solution to many problems often faced by traditional methods e.g. Neural networks are particularly effective in modeling complex patterns and relationships in transformer data. Deep learning variants of neural networks have shown promise in fault diagnosis and anomaly detection. Similarly, SVMs are useful in particular when the data has clear margins of separation. Other techniques like decision tree and random forests provide insights into the importance of different features, which is valuable for understanding the factors influencing transformer health. In particular, traditional techniques have laid a solid foundation, the capabilities of machine learning in handling complex, real-time data make it a superior choice for modern power systems. By integrating ML models, we can achieve more accurate, timely, and reliable assessments, ultimately extending the operational life of power transformers and ensuring the stability of power grids. The future of transformer health assessment lies in leveraging these advanced technologies to meet the growing demands and complexities of modern electrical infrastructure. Some insight into the future trends has been provided below.

The future trends of applying ML to power transformers can be categorized into two primary domains. A notable challenge in harnessing ML lies in the limited availability of failure data during the training phase. Given that power transformers predominantly operate under normal conditions, the data collected during this period often lacks the diversity necessary for effective ML. ML algorithms necessitate exposure to frequently occurring disturbances or faults, which are typically scarce compared to the abundantly available datasets representing normal conditions. Consequently, substantial efforts are imperative to develop robust models capable of generating the requisite dataset for the training phase of ML approaches. Furthermore, power transformers exhibit variances across various facets, including insulating oil volume, construction, insulation materials, environmental conditions, and voltage classes during operation. Coupled with the inherent uncertainty associated with predicting states in oil-filled power transformers, achieving entirely error-free predictions becomes an impractical endeavor, despite the potential for algorithms with higher accuracy. The ensuing objective revolves around surmounting challenges such as imbalanced datasets, real-time fault detection, the intricacy of on-field condition assessment, and the localization capabilities of intelligent systems for transformer health assessment. These are among the pivotal areas necessitating further research and incorporation into future endeavors, in order to propel the application of ML in this domain. Future work is discussed in Section suggesting possibilities of use of ML in addressing challenges such as limited failure data, dataset diversity, transformer variations, imbalanced datasets, real-time fault detection, on-field condition assessment intricacies, and localization capabilities for transformer health assessment.

## 6. Conclusion

This paper discusses the importance of Health Index/condition monitoring for power system equipment, particularly transformers, and the role of ML in analyzing transformer conditions. The conventional methods for diagnosing transformer conditions are reviewed, highlighting their limitations and challenges. Traditionally, offline detection and prediction methods are being used to assess health index and internal state of transformer oil paper insulation, such as BDV test, various spectroscopy methods, DGA, and acidity measurement for a reference OQIN. This review suggests that OQIN and TAN are two characterization techniques that more accurately reflect transformer oil aging compared to other methods that may not directly correlate with aging. These conventional methods used in transformer diagnostic programs were reviewed, and the challenges associated with them were addressed. This paper also discusses the need and importance for transformer insulating oil, which gives various functions like insulation, arc extinguishing and cooling. Research has led to the exploration of alternative insulating oils, however, transformer oil aging can still occur, affecting insulation characteristics and potentially leading to economic and fatal consequences if left undetected or unattended. This discussion emphasizes how ML-based methods offer solutions to these challenges.

In addition to conventional diagnostic techniques, paper focuses and highlight the incorporation of AI, particularly neural networks, to improve the accuracy of condition assessment and life prediction for oil-immersed power transformers. This study provides an extensive overview of present and future functions and uses of ML approaches in assessing transformer conditions. ML has the potential to bring about a fundamental paradigm shift, particularly in situations where traditional model-based and analytical approaches struggle to operate effectively with large volumes of data exhibiting diverse spatial and temporal characteristics. In such cases, ML intelligence excels at generating data-based predictions and judgments based solely on input data, offering valuable insights and opportunities to model unknown processes that were previously challenging to analyze. Moreover, this research includes a comprehensive review of health assessment techniques using the recent literature, providing other researchers with a deeper understanding of the development process for transformer condition assessment.

**Acronyms**

| | |
|---|---|
| AE | Autoencoder |
| AI | Artificial Intelligence |
| ANN | Artificial Neural Networks |
| BDV | Breakdown Voltage |
| BN | Bayesian Network |
| BPNN | Back-Propagation Neural Network |
| BSS | Blind Signal Separation |
| CM | Condition Monitoring |
| CNN | Convolutional Neural Network |
| DDF | Dielectric Dissipation Factor |
| DDP | Decayed Dissolved Particles |
| DGA | Dissolved Gas Analysis |
| DP | Degree of polymerization |
| DT | Decision Trees |
| DTCM | Diagnostic Testing and Condition Monitoring |
| EMD | Empirical Mode Decomposition |
| FBG | Fiber Bragg Grating |
| FDS | Frequency Domain Spectroscopy |
| FFA | Furfur Aldehyde |
| FRA | Frequency Response Analysis |
| FTIR | Fourier-transform Infrared spectroscopy |
| GA | Genetic Algorithm |
| GMM | Gaussian Mixture Model |
| GPR | Gaussian process regression |
| HI | Health Index |
| HPLC | High-Performance Liquid Chromatography |
| IFT | Interfacial Tension |
| IMF | Intrinsic Mode Function |
| IoT | Internet of things |
| IR | Infrared |
| ITD | Intrinsic Time-Scale Decomposition |
| KNN | K-Nearest Neighbour |
| KPCA | Kernel Principal Component Analysis |
| LOL | Loss-of-Life |
| LTOT | Low Temperature Over Temperature |
| ML | Machine Learning |
| MLP | Multi-Layer Perception |
| MLR | Multiple Linear Regression |
| NN | Neutralization Number |
| ODAN | Oil-directed and Air-natural |
| OFS | Optical fiber sensors |
| OQA | Oil Quality Assessment |
| OQIN | Oil Quality Index Number |
| PD | Partial discharge |
| PL | Photoluminescence |
| PSO | Particle Swarm Optimization |
| RF | Random Forest |
| RS | Rough Set Theory |
| RUL | Remaining Useful Life |
| SAR | Structure-Activity Relationship |
| SRC | Sparse Representation Classifier |
| SVDD | Support Vector Data Description |
| SVM | Support Vector Machine |

| SVR | Support Vector Regression |
| TAM | Transformer Asset Management |
| TAN | Total Acid Number |
| TDLAS | Tunable Diode Laser Absorption Spectroscopy |
| UV–VIS | Ultraviolet-Visible Spectroscopy |

**CRediT authorship contribution statement**

**Syeda Tahreem Zahra:** Conceptualization, Methodology, Writing – original draft. **Syed Kashif Imdad:** Supervision, Validation, Writing – review & editing. **Sohail Khan:** Supervision, Writing – review & editing. **Sohail Khalid:** Writing – review & editing. **Nauman Anwar Baig:** Writing – review & editing, Funding acquisition.

**Declaration of competing interest**

The authors declare that they have no known competing financial interests or personal relationships that could have appeared to influence the work reported in this paper.

**Acknowledgments**

We acknowledge the funding provided by RGU to publish this paper.

**Data availability**

No data was used for the research described in the article.

S.T. Zahra et al.Engineering Applications of Artificial Intelligence 139 (2025) 109474